\documentclass[a4paper, 10pt, conference]{ieeeconf}
\usepackage[left=0.75in, right=0.75in, top=1.444in, bottom=0.75in]{geometry}

\usepackage{xcolor}
\usepackage{lipsum}
\usepackage{cite}
\usepackage{amsmath,amssymb,amsfonts}
\usepackage{algorithmic}
\usepackage{graphicx}
\usepackage{siunitx}
\usepackage[utf8]{inputenc}
\usepackage{textcomp}
\usepackage{amsmath}
\usepackage{amsfonts}
\usepackage{commath}
\usepackage{float}

\usepackage{epsfig}
\usepackage{epstopdf}
\usepackage{amsmath}
\usepackage[fleqn]{nccmath}
\usepackage[linesnumbered, ruled, vlined]{algorithm2e}

\def\BibTeX{{\rm B\kern-.05em{\sc i\kern-.025em b}\kern-.08em
    T\kern-.1667em\lower.7ex\hbox{E}\kern-.125emX}}

\newcommand*{\rom}[1]{\expandafter\@slowromancap\romannumeral #1@}
\makeatother

\SetCommentSty{mycommfont}
\newcolumntype{P}[1]{>{\centering\arraybackslash}p{#1}}
\newcolumntype{M}[1]{>{\centering\arraybackslash}m{#1}}

\IEEEoverridecommandlockouts
\begin{document}

\raggedbottom
\title{\LARGE \bf Real-to-Sim: Predicting Residual Errors of Robotic Systems with Sparse Data using a Learning-based Unscented Kalman Filter}

\author{Alexander Schperberg$^{1}$, Yusuke Tanaka$^{1}$, Feng Xu$^{1}$, \\ Marcel Menner$^{2}$, and Dennis Hong$^{1}$% <-this % stops a space 
\thanks{$^{1}$A. Schperberg, Y. Tanaka, F. Xu, and D. Hong are with the Robotics and Mechanisms Laboratory, Department of Mechanical and Aerospace Engineering, University of California, Los Angeles, CA, USA 90095 {\tt\small \{aschperberg28, yusuketanaka, xufengmax, dennishong\}@g.ucla.edu}}
\thanks{$^{2}$M. Menner is with Mitsubishi Electric Research Laboratories (MERL), Cambridge, MA, 02139, USA. {\tt\small menner@ieee.org}}
\thanks{This work was supported by a grant (N00014-15-1-2064)  from  the  Office of Naval Research (ONR)}
}

\maketitle
%\copyrightnotice

\global\csname @topnum\endcsname 0
\global\csname @botnum\endcsname 0
%============================================================
% Abstract and Keywords
%============================================================
\begin{abstract}
Achieving highly accurate dynamic or simulator models that are close to the real robot can facilitate model-based controls (e.g., model predictive control or linear-quadradic regulators), model-based trajectory planning (e.g., trajectory optimization), and decrease the amount of learning time necessary for reinforcement learning methods. Thus, the objective of this work is to learn the residual errors between a dynamic and/or simulator model and the real robot. This is achieved using a neural network, where the parameters of a neural network are updated through an Unscented Kalman Filter (UKF) formulation. Using this method, we model these residual errors with only small amounts of data --- a necessity as we improve the simulator/dynamic model by learning directly from real-world operation. We demonstrate our method on robotic hardware (e.g., manipulator arm, and a wheeled robot), and show that with the learned residual errors, we can further close the reality gap between dynamic models, simulations, and actual hardware. 
\end{abstract}

%\begin{IEEEkeywords}
%Optimization and Optimal Control
%\end{IEEEkeywords}

%============================================================
% Introduction
%============================================================
\section{Introduction}
One of the primary challenges in the field of robotics is to successfully exploit simulation-based results and apply these results to real-world applications. Simulations are convenient as they provide data with low-cost, and can be run safely for long periods of time without risking potentially expensive hardware. However, in almost all cases (even for high-fidelity and expensive simulators), the simulation and reality will differ, i.e., $\lq$reality gap', due to error or mismatch from sensors and actuators that are difficult to correctly model. Because of this reality gap, it typically takes a significant amount of engineering effort to successfully train an agent in simulation and then transfer this learning to hardware. In previous work (see Section \ref{relatedWorks}), there have been numerous approaches to close this reality gap \cite{real_sim_overview}, however, these approaches still require intricate hand-tuning and substantial data collection. 
\begin{figure}[!t]
    \centering
    \includegraphics[width=0.63\columnwidth]{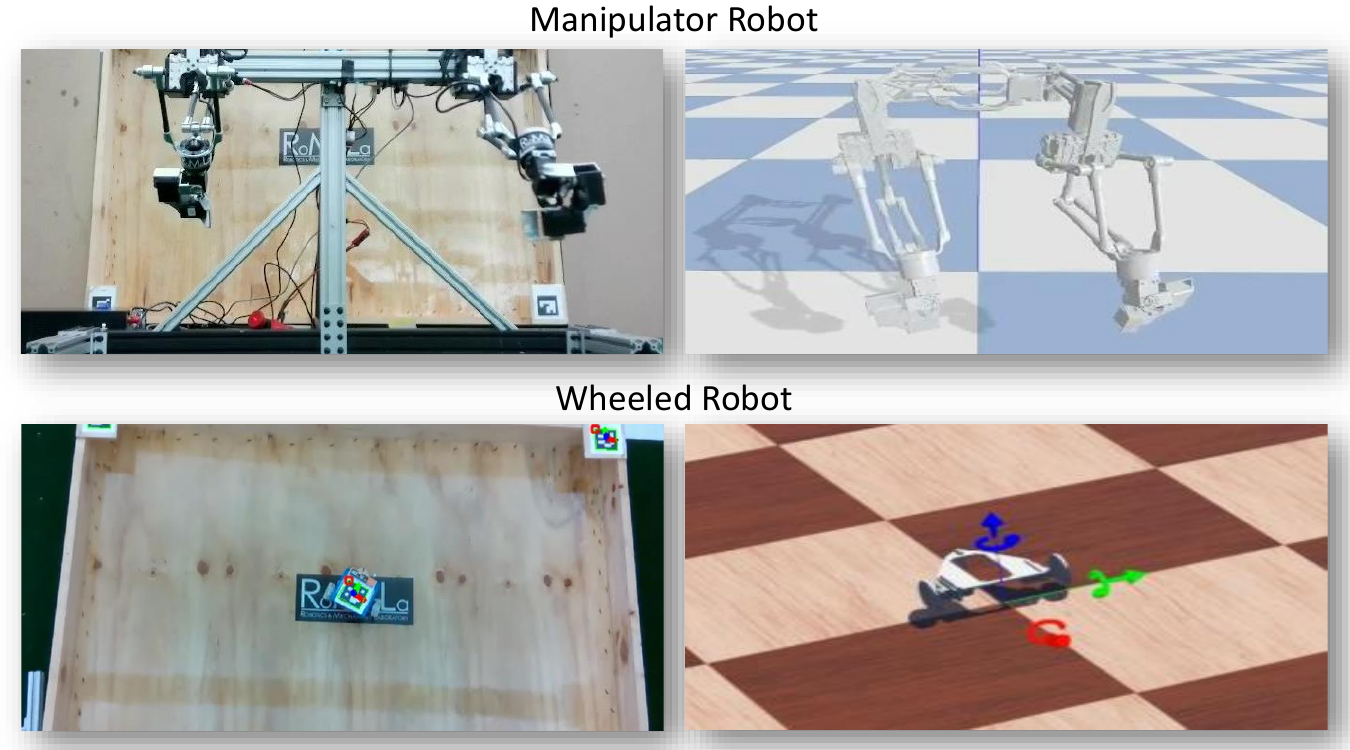}
    \caption{\textbf{Real-to-Sim.} Here we show our experimental setup, where we learn the residual error between the actual and simulated robot.}
    \label{networks}
\end{figure}
Ideally, one way to close the reality gap is to either have an accurate dynamic model of the robot itself and/or directly use hardware data to inform and improve the simulator. Still, the challenge remains in that access to a highly accurate dynamic model is difficult and typically only sparse amounts of data can be received from hardware experiments. Thus, methods that can quickly improve a dynamic model, or a simulator using sparse amounts of data are needed.

In this work, our objective is to provide a method to learn the residual errors assumed to be non-linear and non-Gaussian between a dynamic or simulator model, and the real robot. By learning these residual errors, we can either improve an existing dynamic model to be closer to the real robot, or improve the simulator to be closer to the real robot. The former may be useful for model-based controls (e.g., model predictive control) or for model-based planners, while the latter can be used for reinforcement learning (RL) applications. The residual errors are learned using a neural network, where the parameters of a neural network are updated through an Unscented Kalman Filter (UKF) which can quickly converge to desired parameters (i.e., minimizing the difference between the current and reference model) \cite{menner2021kalman}. A UKF is useful (relative to other filtering methods) as they can better handle non-linear and non-Gaussian residual errors during the update step. 
\subsection*{Summary of our contributions} 
\begin{enumerate}
\item An implementation of neural networks and a UKF to update network parameters are used to quickly learn residual errors between the current and reference model.
\item We show how our approach learns the residual error for several test cases, in particular Sim-to-Dyn (simulator is the reference model and the dynamic model with residual error is the current model), Real-to-Dyn (the real robot is the reference model and the dynamic model with residual error is the current model), and Real-to-Sim (the real robot is the reference model and the simulator with residual error is the current model). 
\item Our results are demonstrated in simulation and hardware using a mobile and stationary manipulator robot.
\end{enumerate}

\section{Related Works}
\label{relatedWorks}
The reality gap between policies trained in simulations and then applied to real-world tasks are typically addressed through domain randomization, domain adaptation, or system identification methods. In domain randomization, a model is trained across simulated environments that vary in its dynamics or visual information. The real-world environment is then assumed to be generalizable as a sub-variant of these randomized environments \cite{auto_sim_real}. However, as sample complexity exponentially increases with the number of randomizations, the work in \cite{Mehta2019ActiveDR} has shown that domain randomization (in general) leads to suboptimal and high variance policies. To resolve the increase of sample complexity over time, \cite{rubrik_cube} addresses this through automatic domain randomization, which expands the range of parameters autonomously during the agent's learning procedure. However, by not using real-world data, their final distribution of parameters may not sufficiently reflect the actual distribution of the real world -- a common issue in this domain \cite{auto_sim_real}.

Another approach is to use real-world data directly to influence simulation parameters. For example, \cite{closing_real_sim_gap} uses continuous object tracking with real-world data to compare trajectories between the real and simulated model. Other works have also employed model-based RL and use real-world data to learn a policy that fits some probabilistic model \cite{ICML2011Deisenroth_323, learning_low_cost}. Still, using machine learning only on real-world data to address the reality gap can be time consuming and impractical depending on the task or environment complexity, as current methods rely on sufficient data collection and may lead to safety issues for real robots during the lengthy training execution.

A hybrid approach (between using real-world and simulated data) exists through domain adaptation methods. \cite{sim_real_adapt} achieves desired manipulation tasks by learning a policy on the real robot through a collection of unlabeled real-world images, which are then employed along with simulated images to adapt the policy (effectively decreasing the number of real-world data required). However, while \cite{sim_real_adapt} does not improve the simulator itself during the learning process, \cite{auto_sim_real} does improve the simulator by matching reality and reduces (over time) the need for real-world data. Although this hybrid approach has been used with success, it requires a careful and at times complex procedure for generating new simulation data, and the agent's policy may struggle to learn if the distribution of the simulation parameters grow too large \cite{sim_real_adapt,learning_task_error}. Thus, while RL has made significant progress in handling the error associated with the real and simulated robot model, resolving real-world data sparsity along with generalizing the training process across environments/robots remains a consistent issue. 

Another approach is to apply methods of system identification and Bayesian optimization, such as in \cite{bayesian_calib} which use prior knowledge of the system. These works typically only account for geometric parameters and do not consider non-linearities that arise from kinematic chains and noise in system propagation (although these methods benefit from being applicable on the real world directly without learning requirements). Further, to achieve the accuracy required in the task-relevant state-space domain, non-geometric errors need to be accounted for including friction, temperature, and compliance which is difficult and potentially infeasible to model without learning-based methods \cite{robot_accuracy_kinematic,sim_real_real_sim}.

In this work, our goal is to make it feasible to train on real-world data directly to continuously improve the simulated model, i.e., Real-to-Sim (or sometimes referred to as modelling a $\lq$digital twin' \cite{sim_real_real_sim}) using a learning-based UKF approach. Other works similar to our goal but different in approach can be found in \cite{auto_learning_model}, where the forward kinematics are modeled by optimizing Denavit–Hartenberg parameters through standard gradient descent (SGD) methods. Our work differs as we are not restricted to robots with only revolute or prismatic joins but is generalizable to any robot type (e.g., wheeled robots). Another work in this space is found in \cite{learning_task_error} which learns the non-parametric model error in addition to the existing (uncalibrated) forward model. Similar to our work, \cite{learning_task_error} applies a trade-off between modelling the non-geometric effects by hand and removing the available kinematic model completely. They achieve this through obtaining accurate forward models in the presence of non-linearities by learning residual errors using Gaussian Process Regression (GPR). However, because these residual errors (or in our case, model mismatch from proprioceptive and exteroceptive noise) is not only non-linear but non-Gaussian, we can make use of neural networks instead \cite{saber,risk-averse}.

Ultimately, our approach is to directly learn the model or residual error between a desired or reference model and the current model. The reference or current model may be composed of a simple dynamic model, the state feedback of the simulator, or the state feedback of the real robot (in our case either a mobile or 6 degrees of freedom manipulator robot). To learn this model error we employ the method described in \cite{menner2021kalman}, which uses a UKF to predict control parameters with user-defined training objectives to evaluate the performance of a closed-loop system online through a recursive implementation. This method has previously shown success for multiple applications, from tuning the cost function weights of an LQR controller for autonomous vehicles \cite{menner2021kalman} to tuning a model predictive controller and trajectory planner of a quadruped robot \cite{alex1}. % and tuning an admittance controller for gasping \cite{alex2}. 
Here, we apply this method for tuning the weights of a neural network toward learning the model error. We use a neural network because we assume this model error is non-linear and non-Gaussian and may potentially learn semantic information that is difficult to account for manually. The motivation of using the method described in \cite{menner2021kalman} (against RL or other adaptive methods for example), is because this approach has been shown to update control parameters with small amounts of data \cite{alex1}, and can handle non-linearity in the data due to the UKF formulation. Additionally, as this approach is applied online and recursively, the method can quickly adapt to model changes or sudden change in the localization result that affects the model error. 

%Go into more details of auto tuning, then link back to previous two paragraphs. 

%============================================================
% Methods
%============================================================
\begin{figure*}[!t]
    \centering
    \includegraphics[width=5.6in]{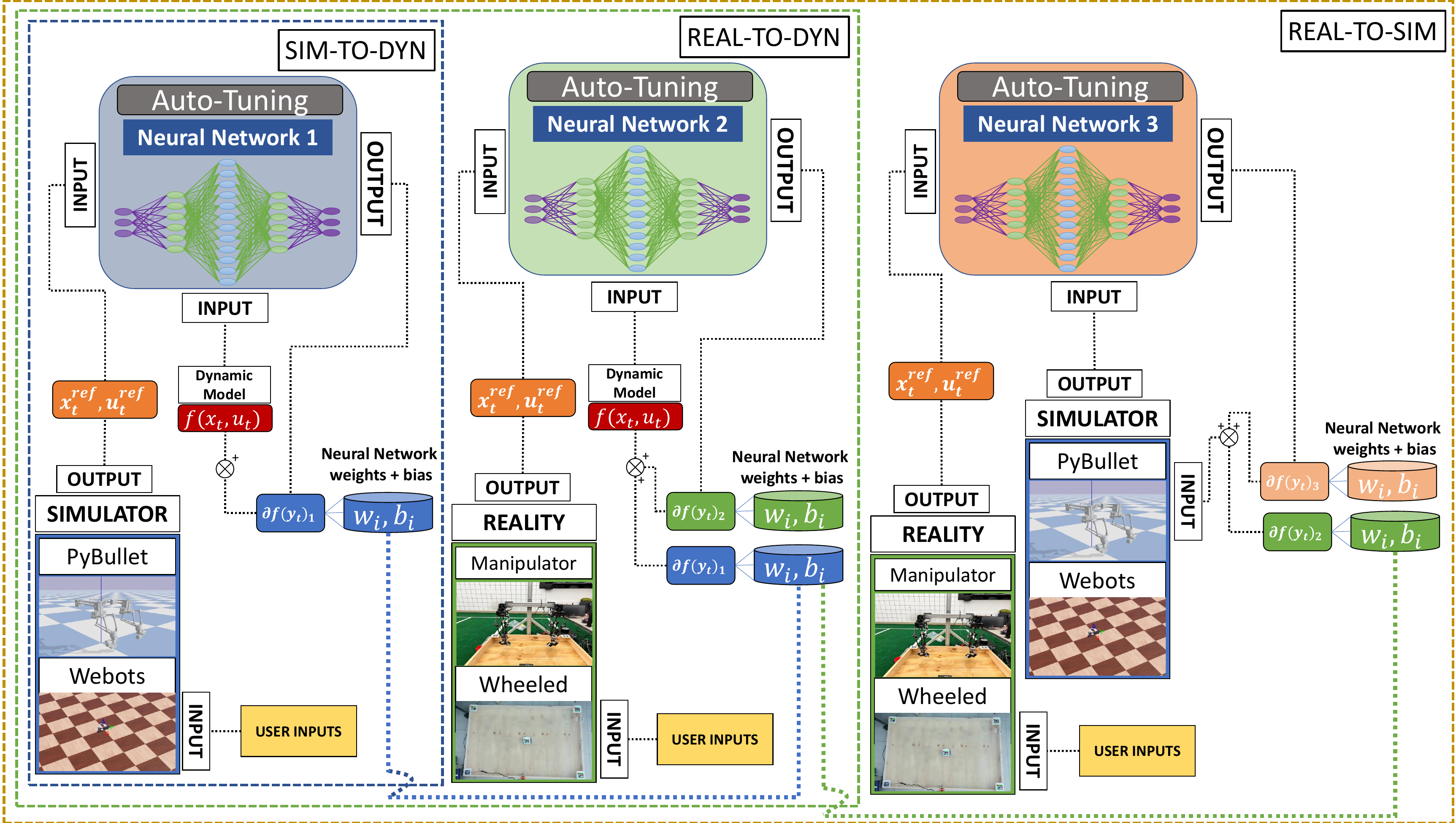}
    \caption{\textbf{Methods flowchart.} Here we show the overall methods of this paper as described in Section \ref{problem_def}. Overall, the goal is to learn the residual model error between a simulator model and a dynamic model or Sim-to-Dyn (boxed in blue), the real robot and a dynamic model or Real-to-Dyn (boxed in green), and the real robot and the simulator model or Real-to-Sim (boxed in brown). The model is learned using a UKF method which calibrates weights and bias parameters of a neural network (the weights and bias variables are parameterized by $\delta(\mathbf{y}_{t})$). Although not necessary, the learned residual model of one case may also be used as part of a warm-start procedure for the next case (e.g., the residual model for Sim-to-Dyn may be used as the starting residual error for the Real-to-Dyn case--which we found to reduce the amount of learning required to model any additional residual error).} 
    \label{fig_methods}
\end{figure*}
\section{Problem Definitions}
\label{problem_def}
\subsection{General Model}
\label{general_model}
We assume the following model propagation from time step $t$ to $t+1$:
\begin{equation}
    \label{generalModel}
    \mathbf{x}_{t+1}=\mathbf{Dyn}(\mathbf{x}_t,\mathbf{u}_t) + \delta(\boldsymbol{y}_t)
\end{equation}
where $\mathbf{x}_{t}$ is the state, $\mathbf{u}_{t}$ is the control input, $\mathbf{Dyn}(\mathbf{x}_{t},\mathbf{u}_{t})$ is some dynamic model of the robot to help propagate the states of the robot from timestep $t$ to $t+1$, and $\delta(\boldsymbol{y}_{t})$ is considered the model mismatch (or residual error of the model), which we assume to be a non-Gaussian and non-linear function parameterized by $\boldsymbol{y}_{t}$. Note, throughout this paper, we will model the derivative terms of the system dynamics since our environment is assumed uniform, and the residual term is not a function of positional space. The objective of this paper is to learn the function $\delta(\boldsymbol{y}_{t})$ through a neural network combined with a UKF (see Section \ref{nn_auto}). Here, we will learn the residual error for several different cases, and transfer the learned residual from one case to the next case to help decrease the overall learning time. Note that from this point, we will use the hat notation on states that result from a predicted model (i.e., dynamic model with the addition of our residual error function) and a bar notation on states that are received directly from the localization result of the simulation or the real robot.

Our first test case will be Sim-to-Dyn, where we use a predefined dynamic model, $\mathbf{Dyn}(\mathbf{x}_t,\mathbf{u}_t)$, and learn the residual error $\delta(\boldsymbol{y}_{t})_{1}$ to predict the state of the model of a simulation, $\mathbf{\hat{{x}}}_{t+1}^{sim}$. Thus, we can write the Sim-to-Dyn case as the following:
\begin{equation}
    \label{simtokin}
    \mathbf{\hat{{x}}}_{t+1}^{sim}=\mathbf{Dyn}(\mathbf{x}_t,\mathbf{u}_t) + \delta(\boldsymbol{y}_{t})_{1}
\end{equation}
The next test case will be Real-to-Dyn, where we try to learn the error between the dynamic model of the robot and the results from the real robot. For this case, instead of starting with just the dynamic model, $\mathbf{Dyn}(\mathbf{x}_{t},\mathbf{u}_{t})$, we instead apply the learned residual from equation $\eqref{simtokin}$ in addition to the dynamic model, and learn the remaining (new) residual error specified by $\delta (\boldsymbol{y}_{t})_{2}$. Note, by using the previously learned model as the updated dynamic model, we may decrease the amount of training time needed to find the remaining residual error (i.e., warm-start) -- however, this is only true if we assume that the simulator model is more accurate than the initialized dynamic model as we do in this paper (although this is not a prerequisite to using our methods and the previously learned model error may also be set to zero if equation \eqref{model_constr} is not true):
\begin{equation}
    \label{model_constr}
    |\Bar{\mathbf{x}}_{t+1}^{real}-\mathbf{Dyn}(\mathbf{x}_{t},\mathbf{u}_t)| > |\Bar{\mathbf{x}}_{t+1}^{real} - {\mathbf{\hat{{x}}}}_{t+1}^{sim}|
\end{equation}
The Real-to-Dyn case can then be specified as:
\begin{equation}
    \label{realtokin}
    \mathbf{\hat{{x}}}_{t+1}^{real}=\mathbf{Dyn}(\mathbf{x}_t,\mathbf{u}_t) + \delta (\boldsymbol{y}_{t})_{1} + \delta(\boldsymbol{y}_{t})_{2}
\end{equation}
where the goal is to learn the residual error, $\delta (\boldsymbol{y}_{t})_{2}$, while $\delta(\boldsymbol{y}_{t})_{1}$ was already learned from the previous Sim-to-Dyn case.

Lastly, we will test the Real-to-Sim case, where our goal is to match the simulator model to the real robot. As before, we can make use of the previously learned cases to inform and facilitate the training procedure. 
For example, we can combine equations \eqref{simtokin} and \eqref{realtokin} to get the following relationship (with the goal of finding the remaining new residual error $\delta f(\boldsymbol{y}_{t})_{3}$):
\begin{equation}
    \mathbf{\hat{{x}}}_{t+1}^{real}=\mathbf{Dyn}(\mathbf{x}_t,\mathbf{u}_t) + \delta (\boldsymbol{y}_{t})_{1} + \delta (\boldsymbol{y}_{t})_{2} + \delta (\boldsymbol{y}_{t})_{3}
\end{equation}
which can be identically represented as: 
\begin{equation}
    \mathbf{\hat{{x}}}_{t+1}^{real}=\mathbf{\hat{{x}}}_{t+1}^{sim} + \delta (\boldsymbol{y}_{t})_{2} + \delta (\boldsymbol{y}_{t})_{3}
\end{equation}
However, we note that in this test case we aim to compare the state estimation of the simulator to the state estimation of the real robot. Thus, we can replace the predicted simulator model, $\mathbf{\hat{{x}}}_{t}^{sim}$, with the actual simulator values defined by $\bar{\mathbf{x}}_{t}^{sim}$ (in this case, $\delta (\mathbf{y}_{t})_{2}$ may act as a $\lq$warm-start' for modeling the behavior of the real robot, since the term was learned during Real-to-Dyn): 
\begin{equation}
    \mathbf{\hat{{x}}}_{t+1}^{real}=\bar{\mathbf{x}}_{t}^{sim} + \delta (\boldsymbol{y}_{t})_{2} + \delta(\boldsymbol{y}_{t})_{3}
\end{equation}
The overall method is also demonstrated in Fig. \ref{fig_methods}. 

\subsection{Neural Network with Unscented Kalman Filtering}
\label{nn_auto}
As described previously, the goal is to learn the residual error $\delta (\boldsymbol{y}_{t})$ from equation \eqref{generalModel} in order to minimize the difference between the current and a reference model. Depending on the test case, what is designated as the current and reference model may differ. For example, in Sim-to-Dyn the reference model is the simulator while the current model is the specified dynamic model, in Real-to-Dyn, the reference model is the real robot while the current model is the specified dynamic model, and in Real-to-Sim, the reference model is the real robot while the current model is the simulator model. Each of these cases and which specific residual model we wish to learn is described in Section \eqref{general_model}. 

\begin{figure}[!t]
    \centering
    \includegraphics[width=1.0\columnwidth]{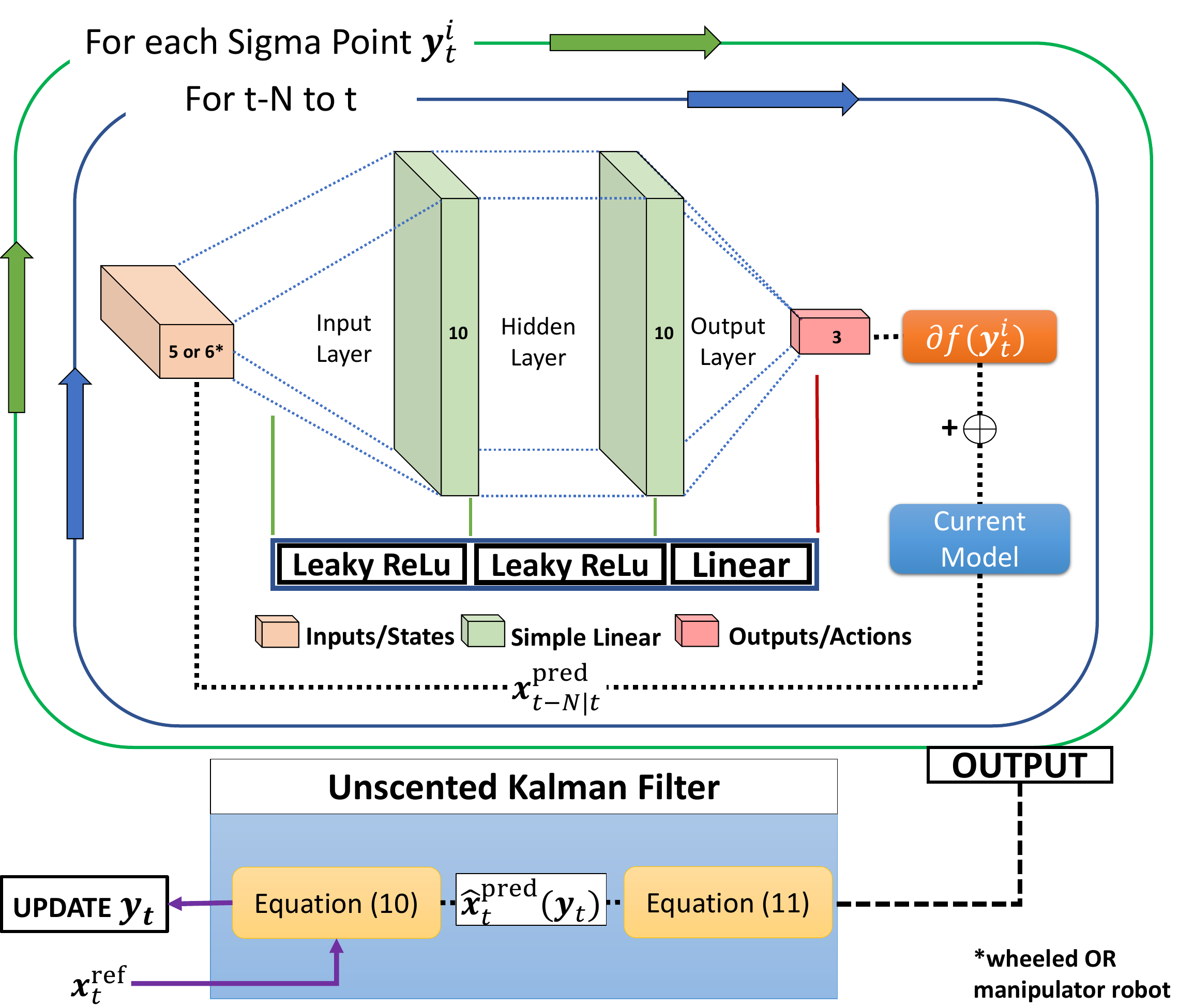}
    \caption{\textbf{Neural Network with UKF.} We illustrate the UKF method as described in Section \ref{nn_auto}. The UKF calibrates weights and bias values of a neural network that is composed of a hidden layer. The result of this calibration is to produce an output or residual model error that minimizes the difference between a current and reference model. The green and blue arrows simply indicate that we are doing a two sequential for loop iterations.}
    \label{networks}
\end{figure}

To learn this residual error $\delta (\boldsymbol{y}_{t})$ we use a fully connected neural network (see Fig. \ref{networks}) that has one input layer (with either 5 inputs when using our mobile robot or 6 inputs when using our manipulator robot), one hidden layer with 10 nodes, and one output layer (with 3 outputs when using our mobile or manipulator robot). We can write this network as the following:
\begin{equation}
\label{nn}
\delta(\boldsymbol{y})=\boldsymbol{y}_{\text {out }} \sigma\left(\boldsymbol{y}_{\text {lay }} \sigma\left(\boldsymbol{y}_{\text {in }}\mathbf{z}+\boldsymbol{y}_{\text {in }, 0}\right)+\boldsymbol{y}_{\text {lay }, 0}\right)+\boldsymbol{y}_{\text {out }, 0}
\end{equation}
where $\boldsymbol{y}_{\rm out}\in\mathbb{R}^{3 \times 10}$ are the weights of the output layer, $\boldsymbol{y}_{\rm out,0}\in\mathbb{R}^{3 \times 1}$ are the bias of the output layer, $\boldsymbol{y}_{\rm lay}\in\mathbb{R}^{10 \times 10}$ are the weights of the hidden layer, $\boldsymbol{y}_{\rm lay,0}\in\mathbb{R}^{10 \times 1}$ are the bias of the hidden layer, $\boldsymbol{y}_{\rm in}\in\mathbb{R}^{10 \times 5}$ or $\in\mathbb{R}^{10 \times 6}$ are the weights of the input layer, and $\boldsymbol{y}_{\rm in,0}\in\mathbb{R}^{10 \times 1}$ are the bias of the input layer, and lastly, the inputs are represented by $\mathbf{z}\in\mathbb{R}^{5 \times 1}$ or $\in\mathbb{R}^{6 \times 1}$. Thus, in total, $\delta(\boldsymbol{y})$ is a function parameterized by $\boldsymbol{y}\in\mathbb{R}^{198 \times 1}$ for the mobile robot, and $\boldsymbol{y}\in\mathbb{R}^{209 \times 1}$ for the robot manipulator. The $\sigma$ represent leaky ReLu activation functions $\mathbf{a}=\sigma(\mathbf{b})$ with $\mathbf{a}=\text{max}(0.01\mathbf{b},\mathbf{b})$, where $\mathbf{b}$ is the input to the activation function. 

Equation \eqref{nn} is considered the feed-forward procedure of a neural network. However, in this paper, instead of using typical back-propagation methods with optimization functions such as Batch Gradient Descent, we instead apply the UKF method described in \cite{menner2021kalman}. The motivation of using this method opposed to other more typical back-propagation methods is that it's model-based and has been shown in previous works \cite{alex1,alex2} to tune parameters quickly with sparse amounts of data. Additionally, compared to gradient-descent methods, a UKF can be more robust to the presence of outliers in the data (i.e., uses a weighted average of sigma points), has reduced computational complexity as it uses a fixed set of sigma point to represent the system state rather than requiring the calculation of higher-order derivatives, and shows improved convergence as it uses deterministic sampling process to propagate the sigma points through the system.

%Because in most cases we have some idea of what the model of our system looks like (e.g., wheeled robots with a differential drive model, or manipulator robot that utilizes forward/inverse kinematics), we can decrease the amount of residual error that needs to be learned by incorporating this model into the training procedure (whereas model-free methods would require large amounts of data to potentially first match a predefined dynamic model before improving upon this model). Decreasing the training time is critical for our application as we seek to learn the residual error between the real robot and a simulator (or kinematic model) using data directly from the real robot (i.e., the amount of data we can collect will be limited). 

While we refer readers to \cite{menner2021kalman} for a full description of the UKF method, the main objective of how we apply this method in this paper, is to update the neural network weights/bias parameters in \eqref{nn} or $\boldsymbol{y}_{t}$ (where $t$ is the current time step) such that the difference between a predicted model, $\mathbf{x}_{t}^{\rm pred}$, and a reference model, $\mathbf{{x}}_{t}^{\rm ref}$, is minimized. This calibration is done through finding a Kalman gain $\mathbf{K}_{t}$ of the UKF using a recursive implementation. In other words, we first collect the history of past reference and predicted values using a time horizon specified by $t-N$ (where $N$ is the number of time steps in the past time horizon), and make the Kalman gain update at the current time step $t$. Thus, we formulate the update of our neural network parameters as:

\begin{align}
\label{eq:learning_law}
    \boldsymbol{y}_{t} &=\boldsymbol{y}_{t-N} + 
    \Delta\boldsymbol{y}_t,
\end{align} 
where the update $\Delta \mathbf{y}_{t}$ is computed based on the following equation:

\begin{align}
\label{eq:UKF}
    \Delta\boldsymbol{y}_t
    =
    \mathbf{K}_t \left(\mathbf{x}^{\rm ref}_{t} - \mathbf{\hat{{x}}}^{\rm pred}(\boldsymbol{y}_{t})\right)
\end{align}
Finally, the Kalman gain ($\mathbf{K}_{t}$) is calculated using the following UKF formulation:

\begin{subequations}
\label{eq:UKF}
\begin{align}
    \mathbf{K}_{t}
    &=
    \mathbf{C}^{sz}_{t}\mathbf{S}_{t}^{-1}
    \\
    \mathbf{S}_{t} 
    &=  
    \textstyle
    {\rm \mathbf{C}}_v + \sum_{i=0}^{2L}\mathbf{w}^{c,i}(\mathbf{x}^{\rm pred,i}_{t}-\boldsymbol{\hat x}^{\rm pred}_{t})\\
    &(\mathbf{x}^{\rm pred,i}_{t}-
    \boldsymbol{\hat x}^{\rm pred}_{t})^{\top}&
    \\
    \mathbf{C}_{t}^{sz}
    &=
    \textstyle
    \sum_{i=0}^{2L}\mathbf{w}^{c,i}(\boldsymbol{y}_{t}^i-\boldsymbol{\hat y}_{t})(\mathbf{x}^{\rm pred,i}_{t}-\boldsymbol{\hat x}^{\rm pred}_{t})^{\top}
    \\
    \boldsymbol{\hat x}^{\rm pred}_{t}
    & = 
    \textstyle
    \sum_{i=0}^{2L}\mathbf{w}^{a,i} \mathbf{x}^{\rm pred,i}_{t}
    \\
    \label{eq:sigmapoints_h}
    \mathbf{x}^{\rm pred,i}_{t}
    & =
    \mathbf{x}^{\rm pred}(\boldsymbol{y}_{t}^i)
    \\
    \mathbf{P}_{t|t-1} 
    &=
    \textstyle
    {\rm \mathbf{C}}_{\theta} + \sum_{i=0}^{2L}\mathbf{w}^{c,i}(\boldsymbol{y}_{t}^i-\boldsymbol{\hat y}_{t})(\boldsymbol{y}_{t}^i-\boldsymbol{\hat y}_{t})^{\top}
    \\
    \boldsymbol{\hat y}_{t}
    & = 
    \textstyle
    \sum_{i=0}^{2L}\mathbf{w}^{a,i} \boldsymbol{y}_{t}^i
    \\
    \mathbf{P}_{t|t} 
    &=
    \mathbf{P}_{t|t-1} 
    - \mathbf{K}_{t} \mathbf{S}_{t} \mathbf{K}_{t}^{\top}
\end{align}
\end{subequations}
where $\boldsymbol{y}_{t}^i$ with $i\!=\!0,...,2L$ are the sigma points, $\mathbf{w}^{c,i}$ and $\mathbf{w}^{a,i}$ are the weights of the sigma points, $\mathbf{C}_{t}^{sz}$ is the cross-covariance matrix, $\mathbf{S}_{t}$ is the innovation covariance, and $\mathbf{P}_{t|t}$ is the estimate covariance. The weights are user defined, and in this paper, we use the same weights as described in \textit{remark 6} of \cite{menner2021kalman}. Lastly, note that the covariance matrix $\mathbf{C}_{\theta}$ (which is initialized by the user) defines the aggressiveness of the update, while $\mathbf{C}_{v}$ defines the $\lq$weight' given to the components of $\mathbf{x}_{t}^{\rm ref}$ \cite{menner2021kalman}. 

%We now describe how we define our reference state $\mathbf{x}^{\rm ref}_{t}$, and predicted state using a model $\mathbf{x}^{\rm pred}_{t}$, as well as the kinematic model for the differential robot and the manipulator robot in section \ref{implementation}.

%============================================================
% Experimental Results
%============================================================
\section{Implementation}
\label{implementation}

\subsection{Computer specifications}
Our method was performed on a laptop with 4 CPU cores (Intel core i7-8850H CPU at 2.60 Ghz) with a Quadro P3200 GPU. We note that the computation time of our method described in Section \ref{nn_auto} was $\approx0.3$ seconds for each update of our neural network parameters, which includes a time horizon of $N=20$. However, the computation time of the UKF update does increase with either a greater time horizon or number of parameters (i.e., larger neural networks). Although we chose a relatively small network in this work (since we only needed a small network to learn the residual errors for our application), larger networks can still be chosen because the computation time of our method is not critical and can be computed (as we do here) on its own CPU core through multi-processing. In other words, the update to our parameters can be performed at any point during the experiment operation and does not need to be updated at any specific frequency. To show the generality of our method, we apply our method on two different robot configurations (both in hardware and in their state localization methods). 

\begin{figure*}[!t]
    \centering
    \includegraphics[width=6.0in]{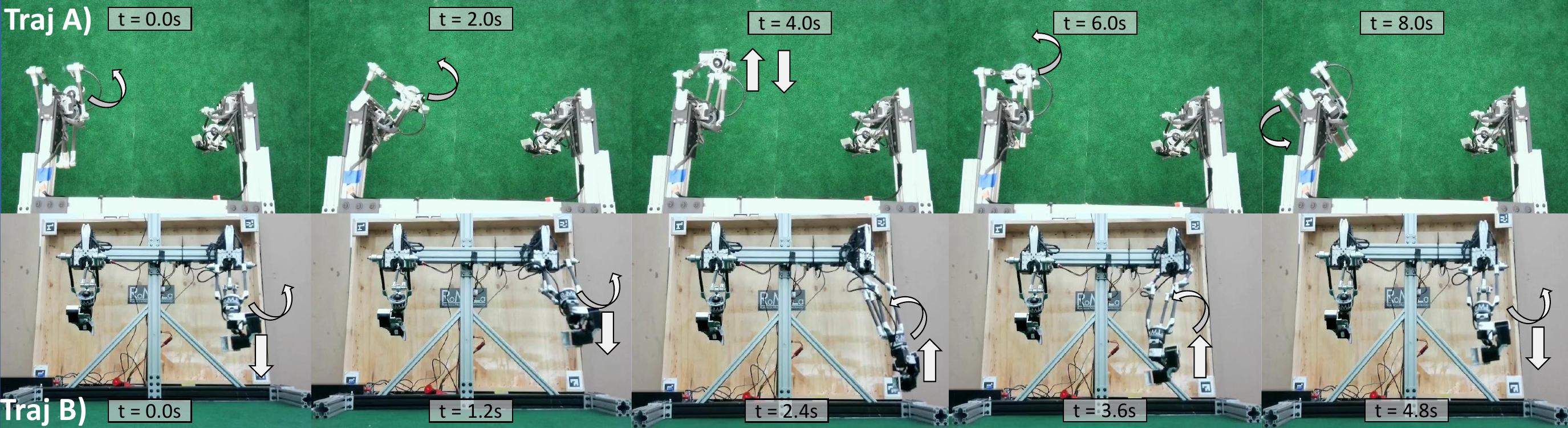}
    \caption{\textbf{Manipulator trajectories.} For evaluating the modelling of residual errors on the manipulator arm, we use two trajectories. The first trajectory (shown in top half) consists of only $x$ and $y$ components--drawing a 2D circle, with a straight line in the middle. This was done to ensure we can model circular and also linear motions simultaneously. The second trajectory, shown on the bottom half, consists of $x$, $y$, and $z$ components--drawing a circle in $x$ and $y$ while moving up and down in $z$. The red line on the last image of the trajectory shows the complete trajectory made by the arm.} 
    \label{traj}
\end{figure*}

\subsection{Differential Drive Robot}
We first demonstrate our methods on a differential drive two-wheeled robot as validation (before applying our methods on our manipulator robot). Localization for the real robot is done using April tags on each corner of a 1 by 1 meter square box and on the robot itself, with an Intel RealSense D435i RGB-D camera in a bird's eye view configuration. The localization provides state estimation of the robot's position and heading angle in addition to their corresponding velocities. For simulating our robot, we use the Webots software \cite{Webots04} which includes system noise and contact dynamics.

To propagate our states (where the residual error is an error on velocity), we use the following differential drive model:
\label{diffRobot}
\begin{equation}
\label{wheel_prop}
\begin{bmatrix}
{\dot{x}}_{t}
\\
{\dot{y}}_{t}
\\
{\dot{\Theta}}_{t}
\end{bmatrix}
= R
\left[\begin{array}{cc}
{}\rm \frac{cos(\Theta_{t})}{2} & {}\rm \frac{cos(\Theta_{t})}{2} \\
{}\rm \frac{sin(\Theta_{t})}{2} & {}\rm \frac{sin(\Theta_{t})}{2}  \\
-\frac{1}{L} & \frac{1}{L} 
\end{array}\right]\mathbf{u_{t}}
+
\delta (\boldsymbol{y}_{t})
\end{equation}

where $\dot{x}$, $\dot{y}$, $\dot{\Theta}$ represent the linear and angular velocity ($\Theta$ is yaw heading angle), $R$ is the radius of the wheel, $L$ is the length from the left to the right wheel, and $\mathbf{u}\in\mathbb{R}^{2 \times 1}$ is the control input, wheel angular velocity for the left ($u^{l}_{t}$) and right ($u^{r}_{t}$) wheel. Lastly, $\delta (\boldsymbol{y}_{t})$ represents the residual error predicted by the neural network (see Section \ref{nn_auto}). For the differential drive robot, our input to this neural network is $\mathbf{z}_{t}$=[$\dot{x}_{t},\dot{y}_{t},\dot{\Theta}_{t},u^{l}_{t},u^{r}_{t}]^{\top}$. Note, that our residual error is an error on the velocity. 

To apply the Kalman gain update as described in equation \eqref{eq:learning_law}, we will use the following for the differential drive robot:
\begin{equation}
\label{training_obj}
\boldsymbol{x}^{\rm ref}_{t}=
\begin{bmatrix}
{\dot{x}}_{t-N|t}^{\rm ref}C_{\dot{x}}
\\
{\dot{y}}_{t-N|t}^{\rm ref}C_{\dot{y}}
\\
\dot{{\Theta}}_{t-N|t}^{\rm ref}C_{\dot{\Theta}}
\end{bmatrix},
\mathbf{\hat{x}}^{\rm pred}_{t}=
\begin{bmatrix}
{\dot{x}}_{t-N|t}C_{\dot{x}}
\\
{\dot{y}}_{t-N|t}C_{\dot{y}}
\\
\dot{{\Theta}}_{t-N|t}C_{\dot{\Theta}}
\end{bmatrix}
\end{equation}
where $\mathbf{x}^{\rm ref}_{t}$ is received directly from state estimation of the simulator or the real robot. $\mathbf{\hat{x}}^{\rm pred}_{t}$ are the values received from the UKF equations described in \eqref{eq:UKF}, and utilizes the dynamic model or the simulator (with the appropriate residual error(s) depending on the test case) from equation \eqref{generalModel}. To clarify the notation, and using Real-to-Dyn as the example \eqref{realtokin}, $\mathbf{x}_{t}^{\rm ref}$ would be equivalent to $\bar{\mathbf{x}}_{t}^{\rm real}$, and $\mathbf{\hat{x}}^{\rm pred}_{t}$ equivalent to $\mathbf{\hat{{x}}}_t^{\rm real}$.

Moreover, we also include a cost term, $C_{\dot{x}}$, $C_{\dot{y}}$, $C_{\dot{\theta}}$, on each component of $\mathbf{x}_{t}^{\rm ref}$ and $\mathbf{x}_{t}^{\rm pred}$---these costs can (if desired by the user) bias learning residual errors of certain components over others (e.g., if the robot's reference trajectory is composed of mainly turning in place, putting a higher cost on $\dot{\Theta}$ may be preferable over other components). In this work, we chose a cost of one for all components and test cases.
\begin{figure*}[!t]
    \centering
    \includegraphics[width=6.0in]{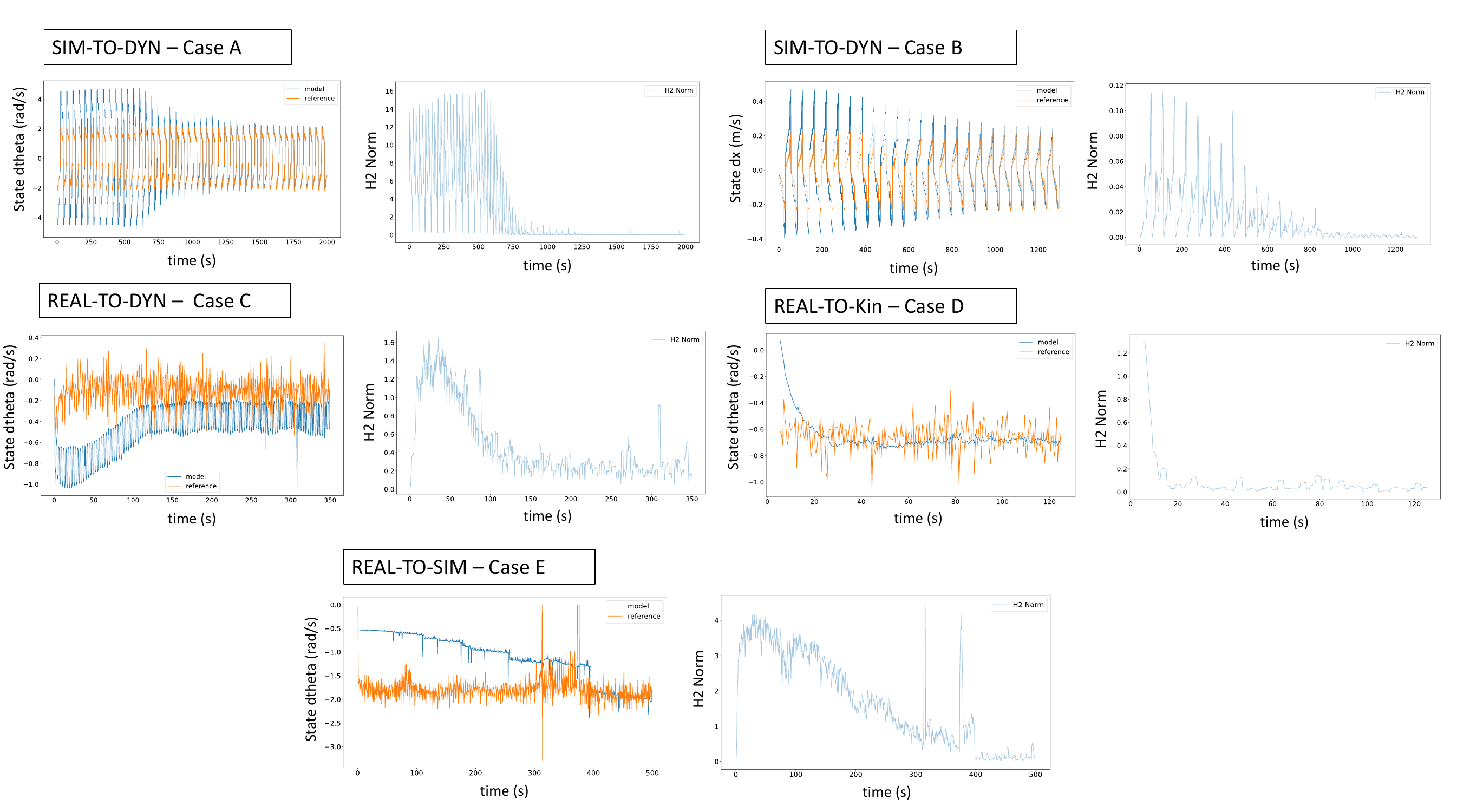}
    \caption{\textbf{Differential drive robot results} The results of learning the residual error of our two-wheeled robot is shown here and described further in Section \ref{results}. In all cases, the H2 norm converges to a steady-state value near zero, indicating the residual model could be learned. We do note that for the real robot (Case C-E) we observed very noisy data due to our hardware (e.g., the wheels would stick and slip on the ground) and error-prone localization. By using a low-pass filter on the output of our neural network however, we could generate a more robust convergence even for this difficult setup.} 
    \label{webots_results}
\end{figure*}

\label{results}
\begin{figure*}[!t]
    \centering
    \includegraphics[width=6.0in]{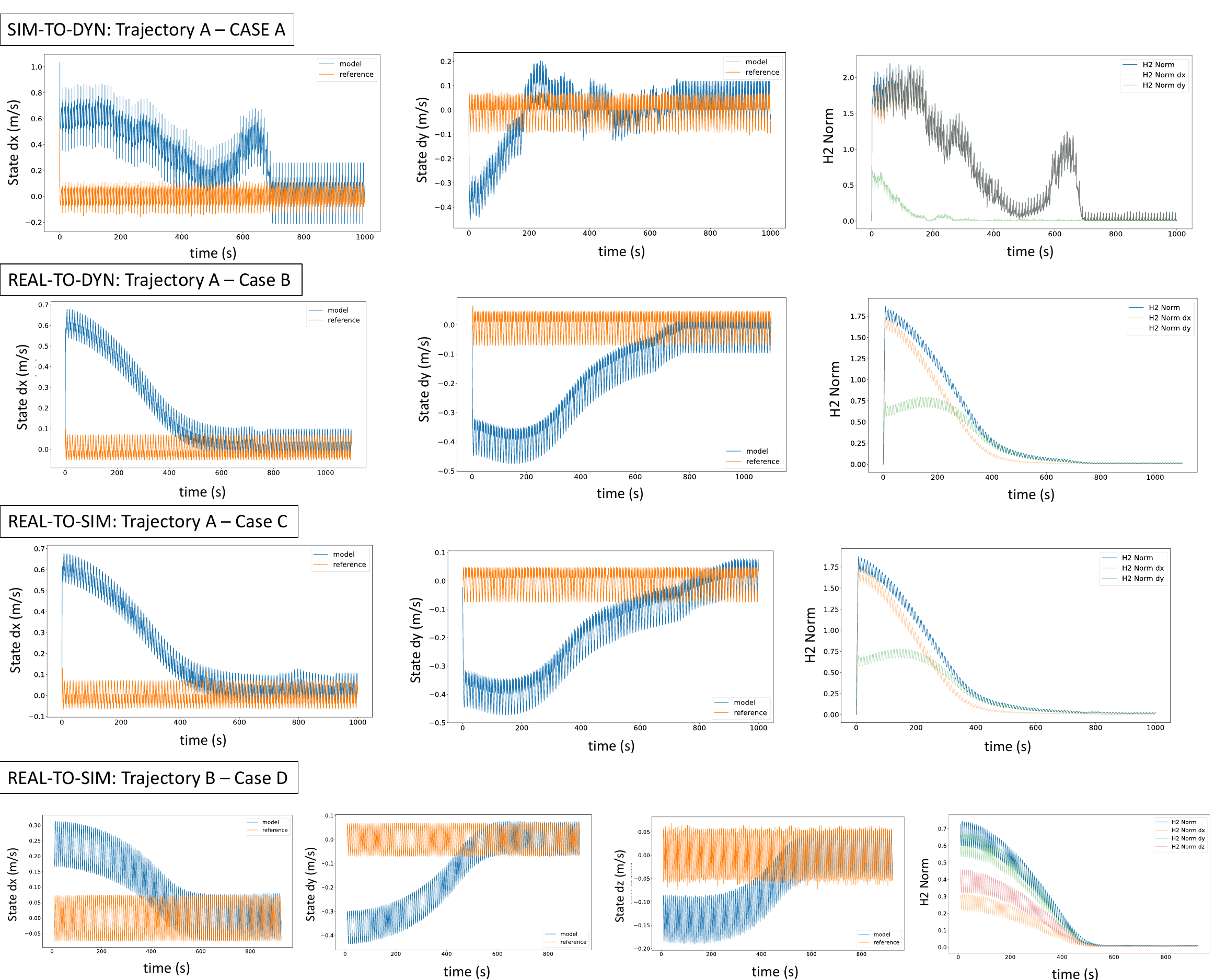}
    \caption{\textbf{Manipulator arm results.} The results of learning the residual error of our manipulator robot is shown here, and described in more detail in Section \ref{results}. For the manipulator arm, we use two trajectories (one consists of 2D motion, i.e., Cases A-C, while another demonstrates 3D motion, i.e., Case D). The motion is described and visualized in Fig. \ref{traj}. Overall, the H2 norm decreased for all cases. We also not only show the overall H2 norm but also the H2 norm of each individual component (i.e., $\dot{x}$,$\dot{y}$, and $\dot{z}$). Unlike the case for our wheeled robot, the localization of our manipulator arm was more stable (we used encoders for localization) and did not require any additional filters to our outputs.}
    \label{scaler_results}
\end{figure*}

\subsection{Manipulator Robot}
\label{ScalerRobot}
After validating our methods on the mobile wheeled robot, we then demonstrate our methods on one of the arms of our manipulator robot SCALER \cite{tanaka2021underactuated}. The arm has 6 degrees of freedom, and is composed of a five-bar linkage combined with a shoulder joint and a spherical wrist joint. Localization for the real robot is done using joint encoders from the motors placed at each joint. For simulating our robot, we use the PyBullet physics engine \cite{benelot2018}. Taking advantage of the rigid body dynamics constraint solver in Pybullet, we simulate the manipulator robot with a closed loop kinematics chain. In the simulator, we use the in-built position control to control the actuators, and set the simulation time step to be 0.01 seconds.  Our objective will be to add the residual error in task space (i.e., on end-effector velocities). The velocities in task space are then used as part of our estimation of $\mathbf{\hat{x}}^{\rm pred}_{t}$. As was done in the implementation of the differential drive robot, we compared $\mathbf{\hat{x}}^{\rm pred}_{t}$ to the reference value $\mathbf{x}^{\rm ref}_{t}$. Thus, to propagate the states of our manipulator robot, we have:
\begin{equation}
\label{fk_prop}
\begin{bmatrix}
    \dot{x}_{t}
    \\
    \dot{y}_{t}
    \\
    \dot{z}_{t}
    \end{bmatrix}
    =\frac{FK (\boldsymbol{\theta}_{t-1}+\dot{\boldsymbol{\theta}}_{t-1}\Delta T)-FK(\boldsymbol{\theta}_{t-1})}{\Delta T}+\delta f(\mathbf{y}_{t-1})
\end{equation}
%\begin{equation}
%    \begin{bmatrix}
%    {x}_{t+1}
%    \\
%    {y}_{t+1}
%    \end{bmatrix}
%    =
%    \begin{bmatrix}
%    {x}_{t}
%    \\
%    {y}_{t}
%    \end{bmatrix}
%    \\
%    +
%    \begin{bmatrix}
%    \dot{x}_{t}
%    \\
%    \dot{y}_{t}
%    \end{bmatrix}\Delta T
%\end{equation}
where $FK(\boldsymbol{\theta}_{t-1})$ represents the forward kinematics of our manipulator robot with joint angles $\boldsymbol{\theta}_{t} \in\mathbb{R}^{6 \times 1}$, joint velocities $\dot{\boldsymbol{\theta}}_{t} \in\mathbb{R}^{6 \times 1}$, and $\dot{x}_{t}$, $\dot{y}_{t}$, and $\dot{z}_{t}$ are the end-effector velocities (a Jacobian can also be used in place of our differentiation). 
The residual error for the manipulator robot, $\delta (\boldsymbol{y}_{t})$, to be predicted by the neural network with the UKF update (see Section \ref{general_model}), will have the following input $\mathbf{z}_{t}=[\dot{\theta}_{t,1},\dot{\theta}_{t,2},\dot{\theta}_{t,3},\dot{\theta}_{t,4},\dot{\theta}_{t,5},\dot{\theta}_{t,6}]^{\top}$, where $\dot{\theta}_{t,1}-\dot{\theta}_{t,3}$ are the shoulder joint velocities, and $\dot{\theta}_{t,4}-\dot{\theta}_{t,6}$ are the spherical joint velocities. Similar to equation \eqref{training_obj} for the differential drive robot, we have the following definitions for $\mathbf{\hat{x}}^{\rm pred}_{t}$ and $\mathbf{{x}}^{\rm ref}_{t}$ (with cost $C=1$ for our case):
\begin{equation}
\label{training_obj2}
\boldsymbol{x}^{\rm ref}_{t}=
\begin{bmatrix}
{\dot{x}}_{t-N|t}^{\rm ref}C_{\dot{x}}
\\
{\dot{y}}_{t-N|t}^{\rm ref}C_{\dot{y}}
\\
{\dot{z}}_{t-N|t}^{\rm ref}C_{\dot{z}}
\\
\end{bmatrix},
\mathbf{\hat{x}}^{\rm pred}_{t}=
\begin{bmatrix}
{\dot{x}}_{t-N|t}C_{\dot{x}}
\\
{\dot{y}}_{t-N|t}C_{\dot{y}}
\\
{\dot{z}}_{t-N|t}C_{\dot{z}}
\\
\end{bmatrix}
\end{equation}

\section{Experimental Results}
Our results are demonstrated for the mobile wheeled robot in Fig. \ref{webots_results} and for our manipulator robot in Fig. \ref{scaler_results}. For both robots, we evaluate the results of our method by comparing the current model (i.e., dynamic or simulator model) with a reference model (simulator model or the real robot) by calculating the H2 norm for each time step during the training process (i.e., defined as $||\mathbf{x}_{t}^{\rm ref}-\mathbf{x}_{t}^{\rm pred}||_{2}$). Thus, if the H2 norm decreases over time and reaches a steady-state value (ideally close to zero), we can assume convergence of the UKF update procedure. In Fig. \ref{webots_results}, we show that the H2 norm decreases for each test case (in light blue) which is estimated based on the model values (dark blue) and reference values (orange). A and B present the Sim-to-Dyn cases, where in A we use a reference trajectory where the robot spins in place and $\dot{\Theta}$ changes from -2 to 2 rad/s, and in B we use a reference trajectory that drives the robot back and forth in the x-direction (where $\dot{x}$ ranges from -0.2 to 0.2 m/s). The Real-to-Dyn cases are shown in C and D (both used a reference trajectory which only changes the angular velocity). However, we note that for the real robot we faced several issues due to localization and hardware. For example, the localization would at times cause large amplitude spikes when estimating the state, and our wheeled robot was made out of low-cost material causing sticking and slipping behavior (as seen by the orange graph in case C). This noise affected the predicted model produced by our UKF update (graph in blue), which still managed to converge but to a local optima solution (as shown by the offset). One option is to introduce a low-pass filter, which we applied on the predicted model output as seen in Case D (we could have also applied the filter on the localization output instead, however, this would not demonstrate as strong of a case for controller robustness under large uncertainty). With the filter, we show a convergence without offset. Lastly, we test the Real-to-Sim case in E (using the filter as done in D) and with a reference trajectory that imposes a constant angular velocity. Some spikes are observed (likely due to bad localization values as seen in the graph in orange) but was able to converge within approximately 17 minutes. 

The results from using our manipulator robot are shown in Fig. \ref{scaler_results}. For cases A - C we used the same reference trajectory as illustrated in the top half or Traj. A of Fig. \ref{traj} (circular 2D motion and drawing a line through the circle---this trajectory was chosen as transitioning from a circular to linear motion causes additional residual error, which we plan to account for with our methods). Note, that the Real-to-Dyn case (or B) produced more robust results (i.e., less transient errors) compared to the Sim-to-Dyn case (or A). One explanation is that as formulated in equation \eqref{realtokin}, the residual model error trained in the Sim-to-Dyn case is used as part of the formulation in equation \eqref{realtokin}. Thus, this additional knowledge may serve as a good initialization for the learning parameters when training the next test case. Finally, we demonstrate two Real-to-Sim cases (C and D), where C is trained on the same reference trajectory as A-B, and D is trained on a reference trajectory shown in the bottom half or Traj. B of Fig. \ref{traj} (A circle for the $x$, and $y$ components and moving up and down in $z$). In both of these cases, the H2 norm reaches near zero and converges. Lastly, we note that convergence typically occurs in approximately 8 minutes of data, and low-pass filtering on model output was not required for our manipulator robot (due to good localization values through joint encoders).

%============================================================
% Discussion
%============================================================
\section{Conclusion}
In this paper we demonstrated a method that can learn and predict the residual model errors between dynamic/simulator models and the real robot. Approximately 17 minutes of experimental data for the wheeled robot and 8 minutes of experimental data for the manipulator robot was required to achieve convergence (i.e., learn the residual model error). Thus, this method is feasible for employment on hardware and with sparse amounts of data.
Although the wheeled robot imposed hardware limitations (i.e., wheels would stick/slide on the surface), and we required low-pass filtering to generate more robust convergence (although convergence was received even without filters), this result showed that experimenting with filtering techniques as part of the parameter calibration process may be promising to increase robustness. One limitation of this work is that the computation time does scale exponentially with an increased neural network. Some analysis in how to reduce the computation time of the algorithm may be needed, so that we can apply our method for more complex tasks (i.e., modelling residual error while grasping an object while considering contact) and legged systems (i.e., quadrupeds and bipeds), and employing larger neural network structures that consists of more challenging data-types (i.e., vision-based data).

%============================================================
% References
%============================================================
\bibliographystyle{IEEEtran}
\bibliography{bibliography}

% *****************************************************************
% 							   END 							    
% *****************************************************************
\end{document}